\documentclass[runningheads]{llncs}
\usepackage[T1]{fontenc}
\usepackage{graphicx}

\usepackage{amssymb}
\usepackage{amsmath}
\usepackage[ruled,linesnumbered]{algorithm2e}
\usepackage{multirow}
\usepackage{multicol}
\begin{document}
\title{Multi-Task Cooperative Learning via \\Searching for Flat Minima}
%
%
\author{Anonymous}
\author{Fuping Wu\inst{1} \and Le Zhang\inst{1} \and Yang Sun\inst{2} \and Yuanhan Mo\inst{2} \and Thomas Nichols\inst{1,2} \and Bart\l omiej W. Papie\.{z}\inst{1,2}}
%
\authorrunning{Anonymous}
%
\institute{Nuffield Department of Population Health, University of Oxford, Oxford, UK \email{\{Fuping.Wu, Le.Zhang\}@ndph.ox.ac.uk} \and
Big Data Institute, University of Oxford, Oxford, UK \\
\email{\{yang.sun, thomas.nichols, bartlomiej.papiez\}@bdi.ox.ac.uk}\\
\email{yuanhan.mo@ndm.ox.ac.uk}
}
\maketitle              

\vspace{-2em}
\begin{abstract}
Multi-task learning (MTL) has shown great potential in medical image analysis, improving the generalizability of the learned features and the performance in individual tasks.
However, most of the work on MTL focuses on either architecture design or gradient manipulation, while in both scenarios, features are learned in a competitive manner.
In this work, we propose to formulate MTL as a multi/bi-level optimization problem, and therefore force features to learn from each task in a cooperative approach.
Specifically, we update the sub-model for each task alternatively taking advantage of the learned sub-models of the other tasks.
To alleviate the negative transfer problem during the optimization, we search for flat minima for the current objective function with regard to features from other tasks.
To demonstrate the effectiveness of the proposed approach, we validate our method on three publicly available datasets.
The proposed method shows the advantage of cooperative learning, and yields promising results when compared with the state-of-the-art MTL approaches.
\textit{The code will be available online.}

\keywords{Multi-Task  \and Cooperative Learning \and Optimization.}
\end{abstract}
\section{Introduction}
With the development of deep learning, multi-task learning (MTL) has shown great potential to improve performance for individual tasks and to learn more transferable features (better generalizability), whilst reducing the number of the network parameters~\cite{ruder2017overview}.
MTL has been widely studied in many domains including image classification \cite{requeima2019fast} or image segmentation \cite{li2022learning}. 
The core assumption behind MTL is that tasks could be correlated and thus provide complementary features for each other \cite{crawshaw2020multi}.
MTL is also applied in medical image analysis tasks \cite{liu2019multi,he2020multi,uslu2021net,he2021hf}, where strong associations between multiple tasks commonly exist. 
For example, the diagnosis of cancer may indicate the extent of disease severity, which can be correlated with the patient's survival, thus diagnosis and prognosis of cancer could be learned simultaneously \cite{shao2020multi}.
In clinical diagnosis, annotations of organs or tissues could support radiologists to grade disease, to mimic this process, Zhou \textit{et.al} \cite{zhou2021multi} studied to simultaneously segment and classify (grade) tumors into benign or malignant class using 3D breast ultrasound images.
Similarly, to improve the prediction of lymph node (LN) metastasis \cite{wang2020ct}, Zhang \textit{et.al} proposed a 3D multi-attention guided multi-task learning network for joint gastric tumor segmentation and LN classification \cite{zhang20213d}.


Typically, MTL methods can be broadly categorized into hard and soft parameter-sharing paradigms \cite{ruder2017overview}.
The former adopts one backbone as the encoder to extract common features for all tasks, and the latter designs encoders for each task while constraining their associated parameters.
To exploit the correlation between tasks, a large amount of work focuses on the architecture design of the network to enable the cross-task interaction \cite{zhang20213d}.
For example, Misra \textit{et.al} designed a cross-stitch model to combine features from multiple networks \cite{misra2016cross}.
Besides network design, many researchers pay more attention to the neural network optimization process to counter the \textit{negative transfer} issue \cite{ruder2017overview}.
As tasks could compete with each other for shared resources, the overall performance might be even poorer than those of solving individual tasks.
To address this issue, previous works either change the weights of each task objective adaptively using heuristics \cite{chen2018gradnorm}, or manipulate the gradient to be descending direction for each task \cite{liu2021conflict}.
However, as those methods formulate MTL in a competitive manner, it is difficult to guarantee that the complementary information is fully utilized by each task.
Moreover, most of them are designed for or evaluated on a simple scenario, where only one domain is involved and the tasks are homogeneous, namely all tasks are either dense prediction or image-level classification.

In this work, we propose a novel cooperative MTL framework (MT-COOL), which manages to update the features of one task while taking into account the current state of other features.
Specifically, we adopt the soft parameter-sharing strategy and update each sub-model conditioning on the information learned by other tasks in an alternative manner.
To avoid the \textit{negative transfer} problem during the training, we further propose to search for flat minima of the current task with regard to others at each iteration.
As a proof of concept, we first validate this method on the simple MNIST dataset for classification tasks. 
To show the advantage of the proposed approach in the medical domain, we use REFUGE2018 dataset for optic cup/disc segmentation and glaucoma classification, and HRF-AV dataset for artery and vein segmentation tasks.
The results show a promising perspective of the proposed multi-task cooperative approach, compared to the state-of-the-art methods.

The main contributions of this work are as follows:
\begin{itemize}\vspace{-0.5em}
    \item We propose a novel MTL framework, which learns features for each task in a cooperative manner.
    \item We propose an effective optimization strategy to alleviate convergence issues.
    \item We validate the proposed method on three MTL scenarios with different task settings.
    The proposed method delivers promising results in all settings, compared with the state-of-the-art MTL approaches.
\end{itemize}


\section{Method}\label{method}
\vspace{-0.7em}
For a better explanation, here we take two-task learning as an example, which can be generalized to n-task problems easily.

\vspace{-0.8em}
\subsection{Bi-Level Optimization for Cooperative Two-Task Learning}\label{method:general}
Formally, let $x_i \in \mathbb{R}^{W\times H \times C}$ denotes an image with the width $W$, height $H$ and channel $C$, 
$y_i \in \mathbb{R}^{C_0}$ is a label for classification, (or $y_i \in \mathbb{R}^{W\times H \times C_0}$ for segmentation) and $C_0$ is the number of classes,
 $F_i(\cdot; \theta_i)$ is a feature extractor,
$ G_i(\cdot; \phi_i)$  is a prediction function for task $i=1,\ldots, T$ where $T$ is a number of tasks, and here $T=2$. $\theta_i$ and $\phi_i$ are corresponding parameters to be learned. 
Our task is to predict label $\widehat{y}_i = G_i(F_i(x_i))$.

For MTL, instead of using shared backbone, \textit{i.e.}, $F_1=F_2$, and updating them simultaneously with a single loss $\ell$, we propose to optimize them in a cooperative manner, that is learning $(F_1, G_1)$ conditioned on a fixed and informative $F_2$, and versa vice.
Generally, it can be formulated as a bi-level optimization problem:
\begin{equation}\label{eq:1}
    (U) \min_{\theta_1, \phi_1} \mathcal{L}_1(\theta_1, \phi_1, \theta_2) = \ell_1(G_1(\mathcal{M}(F_1(x_1; \theta_1),F_2(x_1; \theta_2)); \phi_1), \widehat{y}_1) ,
\end{equation}\\[-1cm]
\begin{equation}\label{eq:2}
    (L) \min_{\theta_2, \phi_2} \mathcal{L}_2(\theta_2, \phi_2, \theta_1) = \ell_2(G_2(\mathcal{M}(F_1(x_2; \theta_1),F_2(x_2; \theta_2)); \phi_2), \widehat{y}_2) ,
\end{equation}
where $\ell_i$ is the loss function, e.g. cross-entropy loss for classification.
$\mathcal{M}$ denotes a feature fusion to facilitate the current task learning by incorporating useful information from other tasks.
A common choice for $\mathcal{M}$ is to use a linear combination of features, also known as \textit{cross-stitch} \cite{misra2016cross} or concatenation operation in multi-layers (which is used in this work due to its simplicity).

To solve the problem Eq.\eqref{eq:1}-\eqref{eq:2}, we propose to update $(\theta_1, \phi_1)$ and $(\theta_2, \phi_2)$ alternatively, as other traditional methods for bi-level optimization problem could be inefficient \cite{biswas2019literature} due to the complexity of deep neural networks.
However, without any constraint, this alternative optimization strategy could fail to achieve convergence to an optimal solution. 
For example, at the $t$-th iteration, we first optimize $\mathcal{L}_1(\theta_1, \phi_1, \theta_2^{(t-1)})$ to obtain an optimum $(\theta_1^{(t)}, \phi_1^{(t)})$.
It is possible that for the second task, $\mathcal{L}_2(\theta_2^{(t-1)}, \phi_2^{(t-1)}, \theta_1^{(t-1)}) < \mathcal{L}_2(\theta_2^{(t-1)}, \phi_2^{(t-1)}, \theta_1^{(t)})$, which means that the update for the first task could increase the prediction risk of the second one, and cancel the gain from optimization of $\mathcal{L}_2$.
Here, we also term this issue as \textit{negative transfer}.
To alleviate this effect, we propose to search for flat minima for one task with regard to the features from the other task in each iteration.

\vspace{-1.2em}
\subsection{Finding Flat minima via Injecting Noise}\label{method:part2}

As mentioned above, the network optimized for one task could be sensitive to the change of parameters for other tasks, which may cause non-convergent solutions.
Hence, at each iteration, for each task, we search for an optimum that is non-sensitive to the update of other parameters within a fixed neighborhood.
We term this kind of optima as \textit{flat minima}.

To formally state this idea, assume that noise $\epsilon_i \sim \{\mathcal{U}(-b, b)\}^{d_{\epsilon_i}}$ with $b>0$, $d_{\epsilon} = d_{\theta_i}$ and $d_{\theta_i}$ the dimension of $\theta_i$.
Then for \textit{task 1}, at $t$-th iteration our target is to minimize the expected loss function with regard to the parameters $(\theta_1, \phi_1)$ and noise $\epsilon_2$, \textit{i.e.,}\vspace{-0.8em}
\begin{equation}\label{eq:3}
    (U)~  \mathcal{R}^{[t]}_1(\theta_1, \phi_1) = \int_{\mathbb{R}^{d_{\epsilon_2}}}\mathcal{L}_1(\theta_1, \phi_1, \theta^{[t-1]}_2+\epsilon_2)dP(\epsilon_2) = \mathbb{E}[\mathcal{L}_1(\theta_1, \phi_1, \theta^{[t-1]}_2+\epsilon_2)] ,
\end{equation}\\[-0.8cm]
\begin{equation}
    s.t. ~|\theta_1-\theta_1^{[t-1]}|<b , \notag
\end{equation}
where $P(\epsilon_2)$ is the noise distribution, and the solution is denoted as $(\theta_1^{[t]}, \phi_1^{[t]})$.
Similarly, for \textit{task 2}, the loss function is as follows,\vspace{-0.5em}
\begin{equation}\label{eq:4}
    (L)~  \mathcal{R}^{[t]}_2(\theta_2, \phi_2) = \int_{\mathbb{R}^{d_{\epsilon_1}}}\mathcal{L}_2(\theta_2, \phi_2, \theta^{[t]}_1+\epsilon_1)dP(\epsilon_1) = \mathbb{E}[\mathcal{L}_2(\theta_2, \phi_2, \theta^{[t]}_1+\epsilon_1)] ,
\end{equation}\\[-0.8cm]
\begin{equation}
    s.t. ~|\theta_2-\theta_2^{[t-1]}|<b . \notag
\end{equation}\\[-0.8cm]

Note that it is hard to find an ideal flat minimum $(\theta^{[t]}_1, \phi^{[t]}_1)$ for Eq.~\eqref{eq:3}, such that $\mathcal{L}_1(\theta^{[t]}_1, \phi^{[t]}_1, \theta^{[t-1]}_2+\epsilon^{(j_1)}_2)=\mathcal{L}_1(\theta^{[t]}_1, \phi^{[t]}_1, \theta^{[t-1]}_2+\epsilon^{(j_2)}_2)$, $\forall \epsilon^{(j_1)}_2, \epsilon^{(j_2)}_2 \sim P(\epsilon_2)$, and $\mathcal{L}_1(\theta^{[t]}_1, \phi^{[t]}_1, \theta^{[t-1]}_2)<\mathcal{L}_1(\theta^{[t-1]}_1, \phi^{[t-1]}_1, \theta^{[t-1]}_2)$, which satisfies the requirement to avoid the optimization issue (see Sect.~\ref{method:general}).
Hence, our goal is to find an approximately flat minimum to alleviate this issue.
A similar idea has been proposed for continual learning \cite{shi2021overcoming}.
However, our method differs as follows: (1) the flat minimum in \cite{shi2021overcoming} is searched for the current task, while in our work, it is searched with regard to other tasks; (2) Once the flat minimum is found for the first task in a continual learning problem, search region for the remaining tasks is fixed, while in our work, the parameters for each task are only constrained in a single iteration, and search region could change during the optimization.

In practice, it is difficult to minimize the expected loss, we instead minimize its empirical loss for Eq.~\eqref{eq:3} and Eq.~\eqref{eq:4} as follows,\vspace{-0.5em}
\begin{equation}\label{eq:5}
    (U)~  L^{[t]}_1(\theta_1, \phi_1) = \frac{1}{M}\sum_{j=1}^{M} \mathcal{L}_1(\theta_1, \phi_1, \theta_2^{[t-1]}+\epsilon_2^{(j)}) + \lambda\cdot  KL(\widehat{y}_1^{(j)} , \frac{1}{M}\sum_{n=1}^{M}\widehat{y}_1^{(n)}) ,
\end{equation}\\[-0.8cm]
\begin{equation}\label{eq:6}
    (L)~  L^{[t]}_2(\theta_2, \phi_2) = \frac{1}{M}\sum_{j=1}^{M} \mathcal{L}_2(\theta_2, \phi_2, \theta^{[t]}_1+\epsilon_1^{(j)}) + \lambda\cdot KL(\widehat{y}_2^{(j)} , \frac{1}{M}\sum_{n=1}^{M}\widehat{y}_2^{(n)}),
\end{equation}\\[-0.4cm]
where $\epsilon_i^{(j)}$ is a noise vector sampled from $P(\epsilon_i)$, $M$ is the sampling times, and $KL$ is the Kullback-Leibler Divergence.
The first term in Eq.~\eqref{eq:5} or Eq.~\eqref{eq:6} is designed to find a satisfying minimum for the current task, and the second term enforces this minimum to be flat as desired.

\vspace{-0.8em}
\subsubsection{Warm Up the Network.} 
To initialize the parameters for  Eq.\eqref{eq:3}) and Eq.\eqref{eq:4} with non-sensitive $(\theta^{[0]}_1, \theta^{[0]}_2)$, we minimize the following loss function,\vspace{-0.8em}
\begin{equation}\label{eq:7}
    \mathcal{L}_{total} = \frac{1}{M}\sum_{j=1}^{M}(\mathcal{L}_1(\theta_1+\epsilon_1^{(j)}, \phi_1, \theta_2+\epsilon_2^{(j)})+\mathcal{L}_2(\theta_2+\epsilon_2^{(j)}, \phi_2, \theta_1+\epsilon_1^{(j)})).
\end{equation}\\[-1.4cm]

\subsubsection{Algorithm.} 
We term the proposed \textbf{m}ulti-\textbf{t}ask \textbf{coo}perative \textbf{l}earning method as MT-COOL.
The algorithm is described in Algorithm \ref{algorithm}.
Note that to alleviate the optimization issue discussed in Section \ref{method:general}, after the update for each task, we clamp the parameters to ensure that they fall within the flat region, as described in Line 17 in Algorithm \ref{algorithm}.

\begin{algorithm}[t!]
\scriptsize
\caption{
 Cooperative Learning via Searching Flat Minima
}
\label{algorithm}
\KwIn{Images and labels $(x_i,y_i)$ for task $i\in \mathcal{T}=\{1,2\}$.
Network for both tasks with randomly initialized parameters $\psi_i = (\theta_i, \phi_i)$, $\psi = (\psi_1,\psi_2)$.
Sampling times $M$, inner iteration number $L$, the flat region bound $b$.
The step sizes $\alpha$, $\beta$.
}

\tcc{Warm up the network to obtain initialized parameters $\psi^{[0]}$}
\For{iteration $t=1,2,\cdots, T_w$}{Sampling $\epsilon_i\sim \{\mathcal{U}(-b, b)\}^{d_{\epsilon_i}}$ with $M$ times for $i=1,2$, respectively\;
Compute $\mathcal{L}_{total}$ in Eq. (\ref{eq:7})\;
Update $\psi^{[t]} = \psi^{[t-1]}-\alpha \triangledown\mathcal{L}_{total}(\psi)$\;
}
Start cooperative learning with $\psi^{[0]}=\psi^{[T_w]}$\;
\tcc{Alternative Update $\psi_i$ for task $i=1,2$.}
\For{Outer iteration $t=1,2,\cdots$}{
\For{task $i=1,2$}{
\For{inner iteration $l=1,2,\cdots, L$}
{Sampling $\epsilon_i\sim \{\mathcal{U}(-b, b)\}^{d_{\epsilon_i}}$ with $M$ times for task $i$ \;
Compute $L_{i}^{[t]}(\theta_i,\phi_i)$ in Eq.~\eqref{eq:5} (or Eq.~\eqref{eq:6}) with fixed $\theta_{\mathcal{T}\backslash \{i\}}^{[t-1]}$\;
\eIf{l=1}{
Update $\psi_i^{[t]} = \psi_i^{[t-1]}-\beta \triangledown L_i^{[t]}(\psi_i)$ \;}
{Update $\psi_i^{[t]} = \psi_i^{[t]}-\beta \triangledown L_i^{[t]}(\psi_i)$ \;

}
Clamp $\theta_i^{[t]}$ into $[\theta_i^{[t-1]}-b,\theta_i^{[t-1]}+b]$\;}
}
}
\KwOut{Model parameters $(\theta_1,\phi_1,\theta_2,\phi_2)$.}
\end{algorithm}


\subsubsection{Network Configuration}
\begin{figure}[h!]
\centering
\includegraphics[width=0.9\textwidth]{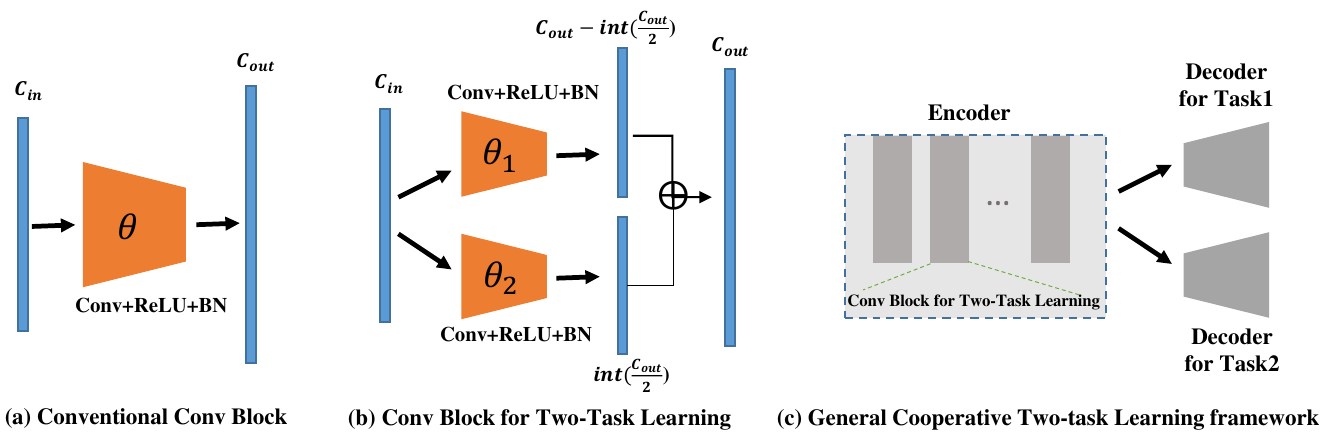}\vspace{-1em}
\caption{A general framework for our MTL method. (a) is the conventional convolution block, (b) illustrates the structure of a convolution block for cooperative two-task learning, and (c) shows the general framework for MTL, which contains an encoder and task-specific decoders.} \vspace{-2em}
\label{fig:1}
\end{figure}

Fig. \ref{fig:1} illustrates the framework for two-task cooperative learning.
Our framework consists of an encoder and task-specific decoders.
The parameters at each layer of the encoder are evenly allocated to each task, and the learned features are then concatenated as the input of the next layer.

\section{Experiments}
We validate our MTL framework in three scenarios as follows: 
(1) classification tasks on different classes with the MNIST dataset \cite{lecun1998gradient}, 
(2) one domain for simultaneous segmentation and classification tasks using the REFUGE2018 dataset \cite{orlando2020refuge}, 
and (3) one domain for two segmentation tasks with HRF-AV dataset \cite{hemelings2019artery}.
For our method, we adopt the stochastic gradient descent (SGD) optimizer, and empirically set the bound value $b=0.05$, the learning rate $\alpha=\beta=0.1$.
To reduce the training time and the memory, we simply set the sampling number $M=1$.
All experiments are implemented using one GTX 1080Ti GPU.

\subsection{Dataset}

(1) \textbf{MNIST.} This dataset contains 50,000 training and 10,000 testing images. 
To simulate a multi-task learning setting, we divide both the training and test images into two subsets with either even numbers $\{0,2,4,6,8\}$ (denoted as \textit{Task 1}) or odd numbers $\{1,3,5,7,9\}$ (denoted as \textit{Task 2}).
For the network, we adopt the widely used LeNet architecture for MNIST dataset \cite{lecun1998gradient}, of which the last layer contains 50 hidden units, followed by a final prediction output. (2) \textbf{REFUGE2018.} The REFUGE2018 challenge \cite{orlando2020refuge} provides 1200 retinal color fundus photography.
The target of this challenge is glaucoma detection and optic disc/cup segmentation.
We divide this dataset into 800 samples for training and 400 test subset, where the ratio of the number of glaucomas to non-glaucoma images are both $1:9$.
As discussed in \cite{orlando2020refuge}, glaucoma is mostly characterized by the optic nerve head area.
Hence, we cropped all images around the optic disc into $512 \times 512$.
We used the UNet \cite{ronneberger2015u} for the segmentation task, with the four down-sampling modules as the shared encoders.
The output of segmentation and the features from the bottom layers are taken as the input of the decoder for classification. (3) \textbf{HRF-AV.} This dataset \cite{hemelings2019artery} contains 45 fundus images with a high resolution of $3504\times 2336$. 
The tasks for this dataset are the binary vessel segmentation and the artery/vein (A/V) segmentation.
We randomly split the dataset into 15 and 30 samples for training and testing.
We adopt the U-Net as the backbone with the bottom feature channel being 256.
During training, we crop patches with size of $2048\times 2048$ randomly as input.

\subsection{Results on MNIST Dataset}

\begin{table}[t]
\centering
\scriptsize
\setlength{\tabcolsep}{6mm}
\renewcommand{\arraystretch}{1.2}
\caption{Performance of SOTA MTL methods on MNIST dataset. We set the number of parameters of \textbf{Joint} method as the base 1, and the values in the column `Params' are the ratio of the parameter number of each method to the \textbf{Joint}.}\label{tab1}
\begin{tabular}{|l|c|l|l|}
\hline
 Methods&  Params & \textit{Task 1}  & \textit{Task 2}\\
 \hline
  Independent& $\approx$ 2   & 99.41 $\pm$ 0.03492  & 98.77 $\pm$ 0.06029 \\
\hline
\hline
Ours (Vanilla) & 1  &99.61$\pm$0.06210    &  99.37$\pm$0.04494 \\

\hline
Ours (\textit{w/o} Reg) & 1  &  99.66$\pm$0.03765   &  99.56$\pm$0.07203 \\

\hline
MT-COOL (Ours) & 1  &  \textbf{99.72$\pm$0.03978}   &  \textbf{99.62$\pm$0.01576} \\

\hline
\hline

Joint & 1  & 99.60  $\pm$ 0.03765 & 99.51 $\pm$0.06281  \\


\hline
CAGrad \cite{liu2021conflict} & 1  &  99.67$\pm$0.05293   &  99.51$\pm$0.05229 \\

\hline
GradDrop \cite{graddrop_chen2020just} & 1  &  99.65$\pm$ 0.03492  &  99.53$\pm$0.04245 \\

\hline
 MGDA \cite{mgd_sener2018multi}& 1  &  99.63$\pm$ 0.05883  &  99.47$\pm$0.05078 \\

\hline
PCGrad \cite{pcgrad_yu2020gradient} & 1  &  99.66$\pm$0.04180   &  99.51$\pm$0.09108 \\
\hline
\end{tabular}
\end{table}

\begin{figure}[!h]
\includegraphics[width=\textwidth]{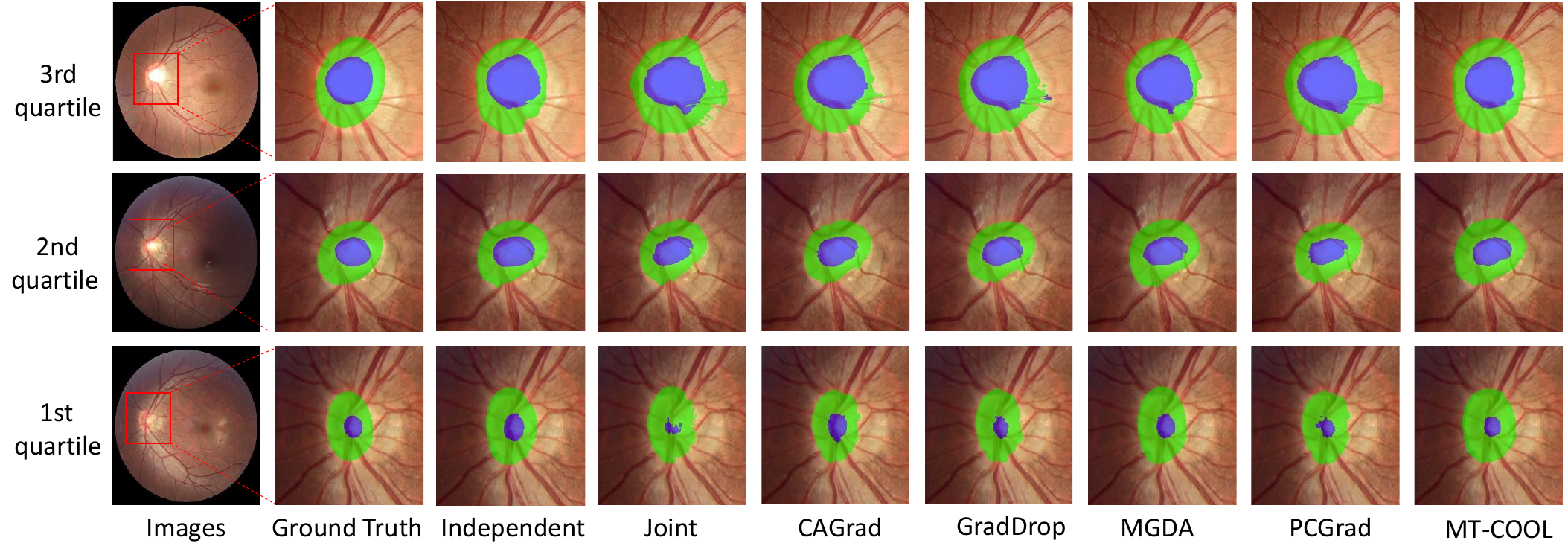}
\caption{Visualization results from MTL methods on REFUGE2018 dataset. The selected samples rank the 1st quartile, median and 3rd quartile in terms of the segmentation performance of \textbf{Independent}.} \label{fig:refuge}
\end{figure}

\begin{table}[t]
\centering
\renewcommand{\arraystretch}{1.5}
\caption{Performance of SOTA MTL methods on REFUGE2018 dataset. }\label{tab:refuge}
\resizebox{1.0\textwidth}{!}{
\begin{tabular}{|l|c|c|c|c|c|c|c|}
			\hline
			\multirow{2}{*}{Methods}&\multirow{2}{*}{Params}& \multicolumn{2}{c|}{Segmentation }& \multicolumn{4}{c|}{Classification} \\
			\cline{3-8}
			&& Cup (Dice\%)& Disc (Dice\%) & Acc  & AUROC & Sen & Spe  \\
			\hline
			\cline{1-8}
			\rule{0pt}{10pt}
			Independent& $\approx$ 2& 95.14$\pm$0.05110  & 86.87$\pm$ 005644  & 0.900$\pm$0.00235 & 0.902$\pm$0.0106  & 0.658$\pm$0.0117  & 0.927$\pm$0.00392  \\
			\hline
   \hline
			Joint& 1& 91.19$\pm$0.7600 &77.36$\pm$0.5236   & 0.907$\pm$0.0183 &0.895$\pm$0.0221   & 0.658$\pm$0.0656  &0.935$\pm$0.0264 \\

			\cline{1-8}
			\rule{0pt}{10pt}

			CAGrad \cite{liu2021conflict}&1& 92.67$\pm$0.7702 & 81.71$\pm$0.2874  & 0.914$\pm$0.00513 &  0.904$\pm$0.00562 & 0.658$\pm$0.0235  & 0.942$\pm$0.00796  \\
   \cline{1-8}
			\rule{0pt}{10pt}
			GradDrop \cite{graddrop_chen2020just}&1&91.70$\pm$0.6376  & 78.91$\pm$1.439 &0.909$\pm$0.00424 & 0.922$\pm$0.0115  & 0.716$\pm$0.0471 & 0.930$\pm$0.00988 \\
   \cline{1-8}
			\rule{0pt}{10pt}
			MGDA \cite{mgd_sener2018multi}&1& 93.87$\pm$0.5017 & 83.87$\pm$0.9732  &  0.895$\pm$0.0154 &0.914$\pm$0.00610  & 0.633$\pm$0.0824  & 0.924$\pm$0.0260\\
   \cline{1-8}
			\rule{0pt}{10pt}
			PCGrad \cite{pcgrad_yu2020gradient}&1&91.74$\pm$0.5569  & 79.80$\pm$0.8748  & 0.911$\pm$0.00849 & 0.898$\pm$0.0136 &0.675$\pm$0.0204  & 0.937$\pm$0.00796 \\
      \cline{1-8}
			\rule{0pt}{10pt}
			MT-COOL (Ours)&1& \textbf{94.37$\pm$0.1706} & \textbf{86.18$\pm$0.3046} & \textbf{0.937$\pm$0.0113} & \textbf{0.942$\pm$0.0149} & \textbf{0.750$\pm$0.000} & \textbf{0.958$\pm$0.0126} \\
			
			\hline

		\end{tabular}
  }
\end{table}

\subsubsection{Ablation Study}
To validate the effectiveness of the two terms in Eq.\eqref{eq:5} and Eq.\eqref{eq:6}, we conduct two experiments: (1) \textbf{Vanilla.} We simply optimize the objective of each task alternatively without any constraints or sampling operations.
(2) \textbf{Ours (\textit{w/o} Reg).} We sample noises during training, and optimize the losses with solely the first term in Eq.\eqref{eq:5} and Eq.\eqref{eq:6}, \textit{i.e.,} without the similarity regularization.
We run 5 times for each method, and report their mean and standard deviation values.

As shown in the top four rows of Table \ref{tab1}, compared to the \textbf{Independent} approach, the proposed \textbf{Vanilla} bi-level optimization method can utilize the features from other tasks and boost the performance of the current one.
By introducing noises to find flat minima during training, \textbf{Ours (\textit{w/o} Reg)} further achieves higher prediction, particularly for \textit{Task 2}.
Finally, by adding similarity regularization, our method obtains the best results.

\subsubsection{Comparison Study}
We compare the proposed method with four state-of-the-art (SOTA) MTL approaches, including MGDA \cite{mgd_sener2018multi}, PCGrad \cite{pcgrad_yu2020gradient}, GradDrop \cite{graddrop_chen2020just} and CAGrad \cite{liu2021conflict}.
We also implement the \textbf{Joint} method as a baseline, which simply sums the loss of each task as the total loss for training.

As shown in Table \ref{tab1}, all MTL methods improve the performance on each task, compared to \textbf{Independent}.
Among all the compared methods, our technique performs the best on both tasks.

\subsection{Comparison on REFUGE2018 Dataset}
For REFUGE2018 dataset, we compare our method with CAGrad, GradDrop, MGDA, PCGrad, and Joint.
We run each method three times, and report the $mean\pm std$ values of Dice score on optic cup and disc for the segmentation task, and accuracy (Acc), Area Under the Receiver Operating Characteristics (AUROC), sensitivity (Sen) and specificity (Spe) for the classification task.

As shown in Table \ref{tab:refuge}, our method achieves comparable results on the segmentation task with the \textbf{Independent}, while other MTL methods degrade significantly, particularly on Disc.
For the classification task, our method achieves the best performance in terms of all the metrics.
Fig. \ref{fig:refuge} provides the visualization results for qualitative comparison.
One can see that the proposed method obtains the best prediction shape among all MTL methods.

\subsection{Comparison on HRF-AV Dataset}

We also conduct a comparison study on HRF-AV dataset.
Each method is repeated three times, and the mean results are presented in Table \ref{tab:hrf-av}.
One can see that compared to the \textbf{Independent}, all the other MTL methods perform poorly, especially on A/V segmentation task.
For example, the best F1 scores on A/V segmentation among the five MTL methods are 0.5127 and 0.5736, respectively, obtained by GradDrop, which are much lower than those from \textbf{Independent}.
On the contrary, our method performs comparably with the \textbf{Independent} on A/V segmentation, and even slightly better on binary segmentation.
For qualitative comparison, please refer to Fig.1 in the Supplementary material.

\begin{table}[t]
\centering
\caption{Performance of SOTA MTL methods on HRF-AV dataset.}\label{tab:hrf-av}
\resizebox{1.0\textwidth}{!}{
\begin{tabular}{|l|c|c|c|c|c|c|c|c|c|}
			\hline
			\multirow{2}{*}{Methods}&\multirow{2}{*}{Params}& \multicolumn{6}{c|}{A/V Segmentation}& \multicolumn{2}{c|}{Binary Segmentation} \\
			\cline{3-10}
			&& Acc (A)& F1 (A) & Acc (V)  & F1 (V) & Acc (AV) & F1 (A/V) & Acc & F1  \\
			\hline
      \cline{1-10}
			\rule{0pt}{10pt}
			Independent&$\approx$ 2& 0.9814 &0.6999   &0.9821  &0.7492   &0.9692   &0.7698   &0.9691   &0.7831    \\

      \hline
      \hline

   Joint&1& 0.9622 & 0.3537  &0.9661  &0.5171   & 0.9664  & 0.7360  & 0.9691  &  0.7835    \\
			\cline{1-10}
			\rule{0pt}{10pt}
   CAGrad \cite{liu2021conflict}&1&0.9687  & 0.4754  &0.9696  & 0.5520  & 0.9668  & 0.7364  &0.9690   &0.7790     \\
			\cline{1-10}
			\rule{0pt}{10pt}
   GradDrop \cite{graddrop_chen2020just}&1&0.9708 & 0.5127  & 0.9716 & 0.5736  &0.9666   &0.7343   &0.9686   & 0.7742     \\
			\cline{1-10}
			\rule{0pt}{10pt}
   MGDA \cite{mgd_sener2018multi}&1&0.9636 & 0.2343  &0.9632  & 0.5315  & 0.9660  & 0.7263  &0.9691   &0.7793    \\
			\cline{1-10}
			\rule{0pt}{10pt}
   PCGrad \cite{pcgrad_yu2020gradient}&1&0.9671  & 0.4262  &0.9681  & 0.5387  & 0.9667  &0.7357   &0.9687   & 0.7763    \\
			\cline{1-10}
			\rule{0pt}{10pt}
			MT-COOL (Ours)&1& \textbf{0.9801} & \textbf{0.6671}  & \textbf{0.9811} &\textbf{0.7135}   & \textbf{0.9674}  & \textbf{0.7424}  &\textbf{0.9701}   & \textbf{0.7912}   \\
			
			\hline

		\end{tabular}
  }
\end{table}

\section{Conclusion}
In this work, we propose a novel MTL framework via bi-level optimization.
Our method learns features for each task in a cooperative manner, instead of competing for resources with each other.
We validate our model on three datasets, and the results prove its great potential in MTL.
However, there are still some issues that need to be studied in the future.
For example, we need to validate our method on large-scale tasks and find a more efficient learning strategy such as using distributed learning.
Moreover, how to allocate the parameters to each task automatically and effectively
is important for model generalization.
For better interpretability, learning features specific to each task should also be studied.



%

%
%
%
\bibliographystyle{splncs04}
\bibliography{reference}
%




\end{document}